%% file: main.tex
\pdfoutput=1

\documentclass[11pt]{article}

\usepackage{acl}

\usepackage{times}
\usepackage{latexsym}

\usepackage[T1]{fontenc}

\usepackage[utf8]{inputenc}

\usepackage{microtype}

\usepackage{inconsolata}

\input{init}

\usepackage{graphics}

\title{\method: Generalizable Multi-aspect Text Evaluation via Augmented Instruction Tuning with Auxiliary Evaluation Aspects}

\author{Minqian Liu$^{\spadesuit}$ \quad \ Ying Shen$^{\spadesuit}$ \quad \ Zhiyang Xu$^{\spadesuit}$ \quad Yixin Cao$^\clubsuit$ \quad \\ \textbf{Eunah Cho}$^\heartsuit$ \quad \textbf{Vaibhav Kumar}$^\heartsuit$ \quad \textbf{Reza Ghanadan}$^\heartsuit$ \quad \ \textbf{Lifu Huang}$^{\spadesuit}$
\\
  $^{\spadesuit}$Virginia Tech \quad $^\heartsuit$Amazon Inc. \quad $^\clubsuit$Fudan University \\ 
  {\tt \{minqianliu, yings, zhiyangx, lifuh\}@vt.edu } \\ {\tt \{eunahch, kvabh, ghanadan\}@amazon.com } \quad {\tt caoyixin2011@gmail.com}
  }

\begin{document}
\maketitle

\input{0abstract}
\input{1introduction}

\input{2relatedwork}
\input{3approach}
\input{4experiment}
\input{5discussion}
\input{6conclusion}
\input{7limitation}

\section*{Acknowledgments}
This research is partially supported by a research award from the Amazon-Virginia Tech Initiative and award No. 2330940 from the Secure and Trustworthy Cyberspace program of the National Science Foundation (NSF). The views and conclusions contained herein are those of the authors and should not be interpreted as necessarily representing the official policies, either expressed or implied, of the U.S. Government. The U.S. Government is authorized to reproduce and distribute reprints for governmental purposes notwithstanding any copyright annotation therein.

\bibliography{custom}

\appendix

\label{sec:appendix}

\input{appendix}

\end{document}

%% file: init.tex
\usepackage{graphicx}               
\usepackage{tabularx}               
\usepackage{booktabs}
\usepackage{amsmath}
\usepackage{stmaryrd}
\usepackage{xcolor}
\usepackage{soul}
\usepackage[linesnumbered,ruled,vlined]{algorithm2e}

\newcolumntype{C}{>{\centering\arraybackslash}X}
\usepackage{multirow}               
\usepackage{diagbox}                
\usepackage{hhline}                 
\usepackage{color, colortbl}                  
\usepackage{amsmath}                
\usepackage{amssymb}                
\usepackage{mathtools}              
\usepackage{enumitem}               

\usepackage{subfigure}
\usepackage{booktabs}
\usepackage{extarrows}
\usepackage{makecell}

\definecolor{RoseQuartzBg}{HTML}{F7CAC9}
\definecolor{RoseQuartz}{HTML}{F5A798}
\definecolor{Serenity}{HTML}{92A8D1}
\definecolor{OrangeRed}{rgb}{1.0, 0.27, 0.0}
\definecolor{Red}{rgb}{1.0, 0.0, 0.0}
\definecolor{Turquoise}{HTML}{0F4C81}
\usepackage{xparse}
\NewDocumentCommand{\lifu}{ mO{} }{\textcolor{OrangeRed}{\textsuperscript{\textit{Lifu}}\textsf{\textbf{\small[#1]}}}}
\NewDocumentCommand{\mo}{ mO{} }{\textcolor{blue}{\textsuperscript{\textit{Mo}}\textsf{\textbf{\small[#1]}}}}
\NewDocumentCommand{\minqian}{ mO{} }{\textcolor{teal}{\textsuperscript{\textit{Minqian}}\textsf{\textbf{\small[#1]}}}}
\NewDocumentCommand{\zhiyang}{ mO{} }{\textcolor{blue}{\textsuperscript{\textit{Zhiyang}}\textsf{\textbf{\small[#1]}}}}
\NewDocumentCommand{\sijia}{ mO{} }{\textcolor{Red}{\textsuperscript{\textit{Sijiia}}\textsf{\textbf{\small[#1]}}}}
\NewDocumentCommand{\ying}{ mO{} }{\textcolor{cyan}{\textsuperscript{\textit{Ying}}\textsf{\textbf{\small[#1]}}}}

\newcommand{\method}{\textsc{X-Eval}}

\newcommand{\evalinstruct}{\textsc{AspectInstruct}}

%% file: 0abstract.tex
\begin{abstract}
Natural Language Generation (NLG) typically involves evaluating the generated text in various aspects (e.g., consistency and naturalness) to obtain a comprehensive assessment.
However, multi-aspect evaluation remains challenging as it may require the evaluator to generalize to any given evaluation aspect even if it's absent during training.
In this paper, we introduce \method{}, a two-stage instruction tuning framework to evaluate text in both seen and unseen aspects customized by end users. 
\method{} consists of two learning stages: the vanilla instruction tuning stage that improves the model's ability to follow evaluation instructions, and an enhanced instruction tuning stage that exploits the connections between fine-grained evaluation aspects to better assess text quality.
To support the training of \method{}, we collect \evalinstruct{}, the first instruction tuning dataset tailored for multi-aspect NLG evaluation spanning 27 diverse evaluation aspects with 65 tasks. To enhance task diversity, we devise an augmentation strategy that converts human rating annotations into diverse forms of NLG evaluation tasks, including \textit{scoring, comparison, ranking}, and \textit{Boolean question answering}.
Extensive experiments across three essential categories of NLG tasks: dialogue generation, summarization, and data-to-text coupled with 21 aspects in meta-evaluation, demonstrate that \method{} enables even a lightweight language model to achieve a comparable if not higher correlation with human judgments compared to the state-of-the-art NLG evaluators like GPT-4.
\footnote{The source code, model checkpoints, and datasets are publicly available at \url{https://github.com/VT-NLP/XEval} for research purposes.}

\end{abstract}


%% file: 1introduction.tex
\section{Introduction}

\input{figures/hint_instance_fig}


Recent advancements of pre-training~\cite{flant5,llama,llama2}, prompting~\cite{GPT3,cot,sccot,tot,socratic}, and instruction tuning~\cite{wei2022finetuned} have improved the quality of machine generated texts by a significant degree. Nevertheless, the evaluation of various Natural Language Generation (NLG) tasks still lags far behind compared with the rapid progress of large language models (LLMs). Previous similarity-based metrics such as ROUGE~\cite{lin2004rouge}, BLUE~\cite{papineni2002bleu}, METEOR~\cite{banerjee2005meteor}, and BERTScore~\cite{Zhang2020BERTScore} predominantly measures the similarity between the generated and reference text, failing to accurately reflect the quality of generated text~\cite{Gehrmannp2023repair}, especially for open-ended generation tasks. 

To obtain a more comprehensive assessment of text quality, multi-aspect evaluation~\cite{fabbri2021summeval} has been proposed to evaluate the generated text from multiple fine-grained evaluation \textit{aspects}, such as \texttt{fluency} and \texttt{consistency}. 
While most existing studies~\cite{mehri2020usr,yuan2021bartscore,unieval} consider a closed set of aspects, in many realistic scenarios, the users may need to evaluate the text with their customized aspects and specifications, calling for building an evaluator that can be flexibly extended to any \textit{unseen} aspects without the need of training data. 
Recent studies~\cite{gptscore, geval} propose to leverage large language models (LLMs) such as GPT-4~\cite{OpenAI2023GPT4TR} as NLG evaluators, yielding promising zero-shot performance on unseen aspects. However, such evaluations, especially with proprietary LLMs, are cost-intensive, time-consuming, and pose concerns about data privacy and reproducibility.




In this work, we propose \method{}, an automatic evaluation framework that can conduct fine-grained evaluation on both seen and unseen aspects across various NLG tasks with a single model, as illustrated in Figure~\ref{fig:teaser}. 
\method{} follows a two-stage training paradigm:  we first instruction-finetune an open-source language model to equip it with the capability of following human-written instructions for evaluation. Then, motivated by the observation that evaluation aspects usually exhibit inter-connections~\cite{gptscore} and thus their evaluations can benefit each other, 
we introduce an additional training stage to finetune the model on the instruction-tuning tasks enriched with the evaluations of a set of \textit{auxiliary aspects}, which are expected to provide clues for evaluating the target aspect and encourage consistent evaluations across multiple aspects. 
During training, for each target aspect, we take all the remaining aspects defined in the corresponding dataset as auxiliary aspects and incorporate their gold evaluations into the instructions for the second-stage tuning.
During inference, given the target aspect, we first select a set of auxiliary aspects based on the similarity of the aspect definitions and predict the evaluation result for each auxiliary aspect using the trained model. We then re-perform the evaluation for each target aspect by incorporating the results of auxiliary aspects.
To support our proposed two-stage training of \method{},
we construct \evalinstruct{}, the first multi-aspect evaluation instruction tuning dataset spanning 27 diverse evaluation aspects over 65 tasks. This dataset is anchored around three core categories of NLG tasks: dialogue, summarization, and data-to-text.
In light of insights from previous studies in instruction tuning~\cite{wei2022finetuned, xu2023multiinstruct}, which emphasize the advantage of task diversity in enhancing zero-short generalization, we further augment the dataset by converting the original human rating data into diverse forms of NLG evaluation tasks, including \textit{scoring, comparison, ranking} and \textit{Boolean question answering}. In addition, to incorporate auxiliary aspects, we manually create templates that convert the numerical evaluation scores of each aspect into descriptions in natural language. 

The main advantages of our approach are highlighted as follows: \textbf{(1) Generalization ability:} we introduce \method{} that can be flexibly generalized to evaluate the unseen NLG tasks or the aspects customized by user instructions in a zero-shot manner with a single model; \textbf{(2) Strong performance with high efficiency:} with significantly less amount of model parameters (780M), \method{} achieves strong performance compared to the state-of-the-art LLM-based evaluators (including GPT-4) demonstrated through comprehensive experiments; \textbf{(3) Reference-free and open-source:} our evaluator does not require gold reference to perform evaluation and it is more reliable and transparent thanks to its open-source nature.


%% file: figures/hint_instance_fig.tex
\begin{figure}[!t]
	\centering
	\includegraphics[width=\linewidth, trim={0 0 0.2cm 0},clip ]{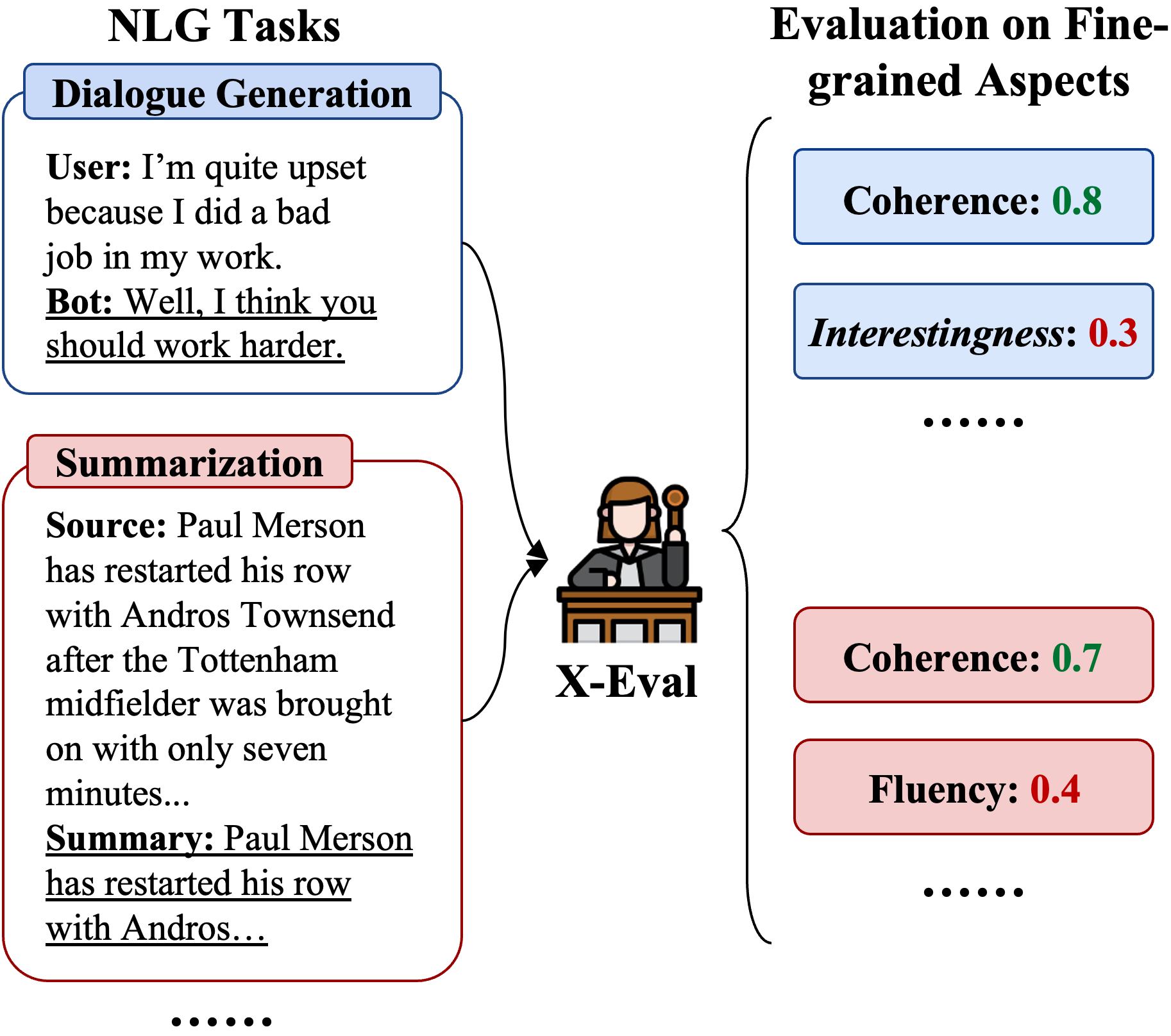}
    \caption{
    Illustration of \method{} for multiple seen and unseen fine-grained evaluation aspects across various NLG tasks. The unseen aspect (i.e., \texttt{Interestingness}) is highlighted in \textit{italics}. The text to be evaluated is highlighted with \underline{underline}. In this example, each evaluation score is from 0 to 1. The higher score indicates better quality.
    }

\label{fig:teaser}
\end{figure}

%% file: 2relatedwork.tex
\section{Related Work}




\paragraph{Similarity-based Metrics}
The previously dominant text evaluation paradigm is to predict a one evaluation score, where most of them are similarity-based metrics, including metrics that measure the surface overlap between the generated and reference text, such as ROUGE~\cite{lin2004rouge}, BLUE~\cite{papineni2002bleu}, and METEOR~\cite{banerjee2005meteor}, as well as metrics measuring the distance between the contextualized embeddings of the generated text and the reference as the similarity score, such as BERTScore~\cite{Zhang2020BERTScore} and MoverScore~\cite{zhao2019moverscore}. Although these metrics are widely adopted, they often overlook fine-grained aspects and later study~\cite{Gehrmannp2023repair} has proven that they fail to truly capture the quality of text with the coarse-grained score. 

\paragraph{Multi-Aspect Metrics} 
To conduct a more holistic evaluation, recent studies~\cite{wang20fact,huang20grade} propose to evaluate the NLG systems via multiple fine-grained aspects. 
UniEval~\cite{unieval} proposes to re-frame NLG evaluation into a QA format and perform multi-aspect evaluation with a single model via continual learning~\cite{madotto-etal-2021-continual,liu-etal-2022-incremental,liu-huang-2023-teamwork}. 
However, UniEval cannot maintain robust performance when generalizing to novel aspects. 
To obtain an evaluator that can be generalized to customized aspects, some recent studies~\cite{gptscore,geval} harness proprietary LLMs to perform fine-grained evaluation in a zero-shot manner. However, due to the closed-source nature, these evaluation metrics suffer from issues of reproducibility and are prohibitively expensive. More recently, some concurrent studies~\cite{instructscore,tigerscore,autoeval,kim2024prometheus} propose to extract instruction-following data from proprietary LLMs for finetuning a more lightweight model as the evaluator. Nevertheless, they still require high costs to call the APIs to obtain a large amount of training data and it is non-trivial to ensure the data are of high quality.
In addition, to the best of our knowledge, we are the first to meticulously curate the instruction-tuning dataset and train an instruction-based evaluator for dialogue evaluation.



%% file: 3approach.tex
\section{\evalinstruct}

\subsection{Problem Definition}

Multi-aspect automatic text evaluation aims to evaluate the quality of NLG system's output $x$ given a set of evaluation aspects $\mathcal{A}$ (e.g., \texttt{coherence}, \texttt{naturalness} and so on), and optionally an additional set of texts $\mathcal{S}$ (e.g., the source documents for text summarization, or context for dialogue evaluation). The evaluation task can be formulated as:
\begin{equation*}
    c = f(x, \mathcal{S}, a)
\end{equation*}
where $a\in\mathcal{A}$ is the fine-grained aspect to be evaluated, and $f(\cdot)$ is the scoring function that provides an assessment $c$ w.r.t. the aspect $a$.

\subsection{Data Collection}
\label{sec:instruction_tasks}

We aim to build a unified automatic evaluation framework that can assess the text quality for both seen and unseen evaluation aspects across various NLG tasks via instruction tuning. 
To this end, we build an instruction-tuning dataset tailored for multi-aspect evaluation, namely \evalinstruct{}, with the following steps:

\paragraph{Existing Dataset Collection} We first collect 10 existing evaluation datasets with human annotations for 3 representative categories of NLG tasks, including dialogue generation~\cite{dailydialog,dstc9,holisticdial,gopalakrishnan2019topical,mehri2020unsupervised}, text summarization~\cite{tldr,fabbri2021summeval,wang2020asking,unieval}, and data-to-text~\cite{wen2015semantically}. 

\begin{figure*}[htbp]
	\centering
	\includegraphics[width=\linewidth, trim={0.1cm 0 0 0},clip ]{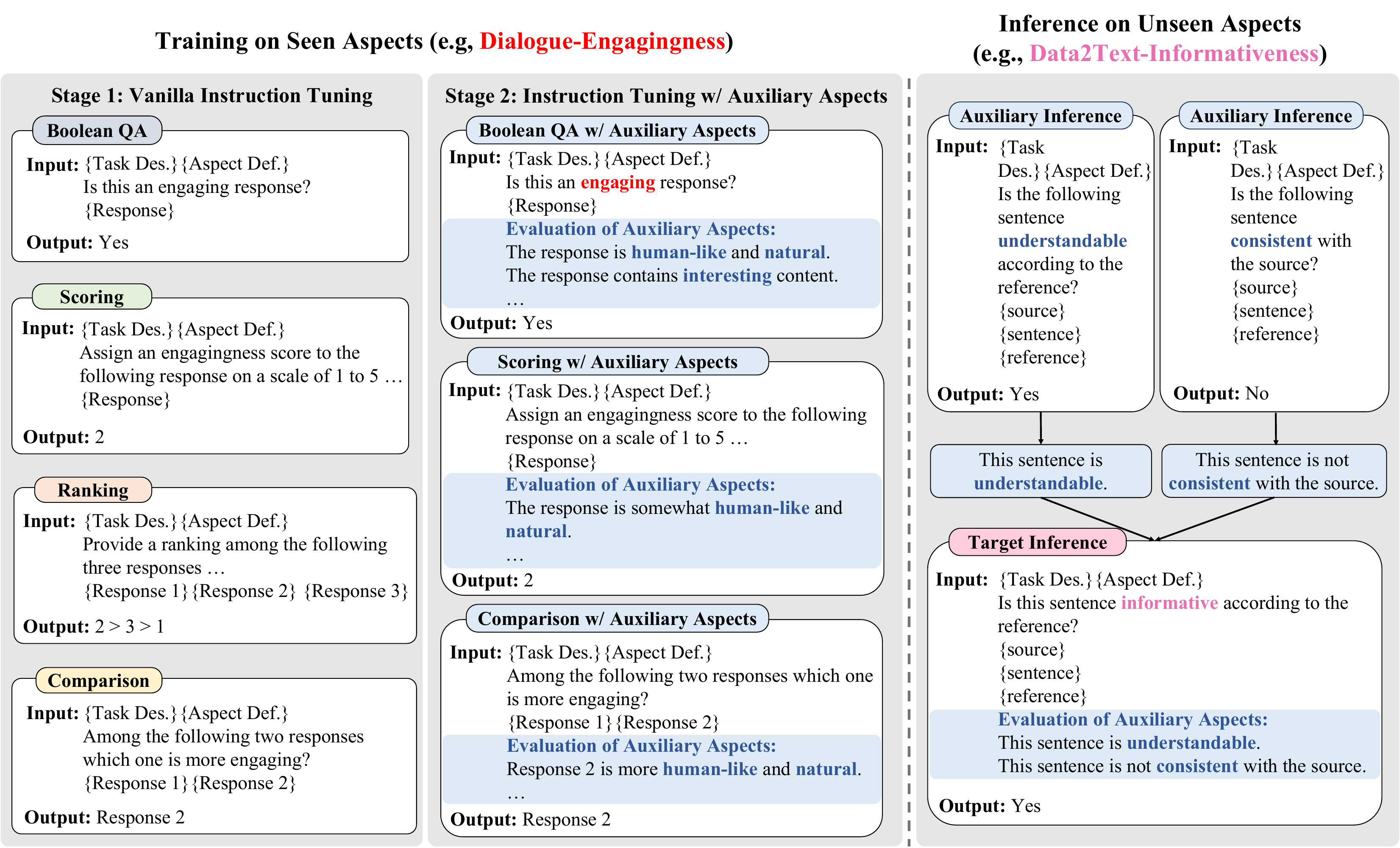}
	\caption{\textbf{Illustration of our \method{} framework.} The left section depicts our two-stage training approach: vanilla instruction tuning on diverse tasks and subsequent training on instruction tasks enriched with auxiliary aspects. The right section illustrates the inference pipeline with auxiliary aspects.
 }
\label{fig_illustrate}
\end{figure*}

\paragraph{Task Augmentation} The original datasets we collect only contain numerical scores annotated by humans, which severely limits the diversity of instruction-tuning tasks. Thus, we further derive diverse forms of evaluation tasks from the original annotations to enhance task diversity. 
Denote the ground truth score for text $x_i$ as $y_i$. We derive four types of tasks based on this annotation: \textbf{(1) Scoring:} we ask the model to directly predict a discrete score (e.g., in the Likert scale) where we map the continuous ground truth $y_i$ into a discrete scale; 
\textbf{(2) Comparison:} we sample two texts $x_i$ and $x_j$ 
for an identical context, e.g., two versions of summaries for the same source document,   
and ask the model to select the text with the higher evaluation score; \textbf{(3) Ranking:} we further extend the comparison task into ranking by sampling three candidates under the same context and ask the model to predict the correct ranking of the candidates based on the text quality; \textbf{(4) Boolean Question Answering:} we also formulate evaluation as a Boolean QA task following \cite{unieval} by asking the model a question such as "\textit{Is this response fluent?}" and let the model predict "\textit{Yes}" or "\textit{No}". 

\paragraph{Instruction Creation} Finally, we define a unified instructions format for tasks included in \evalinstruct{}. Each instruction consists of three parts: (1) \textit{task description} that briefly introduces the evaluation task, (2) \textit{aspect definition}, and (3) \textit{evaluation protocol} that details what the model should output to perform the evaluation.
We present the detailed procedure for instruction annotation in Appendix~\ref{app:annotation}.
We provide an example of the original annotation, and the derived evaluation tasks along with the curated instructions in Figure~\ref{fig_detailed_example} in Appendix~\ref{app:aug_it_task}. 
The full list of evaluation aspects and the collected instructions can be found in Appendix~\ref{app:asp_def}.

\paragraph{Statistics}
In total, we construct 65 tasks in \evalinstruct{}, where we split 32 tasks and 14 seen aspects for instruction tuning and 33 tasks and 13 unseen aspects for meta-evaluation.
We collect 72,637 instances in total with 55,602 instances for training and 17,035 instances for inference. Note that there is no overlap among the datasets used for training and inference. We consider two aspects that have identical aspect names but are in different NLG tasks as distinct aspects. We include more details about the source datasets, constructed instruction-tuning tasks, and the number of instances of each task in Appendix~\ref{app:aug_it_task}.





\section{\method{}}
\label{sec:main_method}
\subsection{Two-Stage Instruction Tuning}

Figure~\ref{fig_illustrate} presents an overview of \method{}, which consists of two stages of instruction tuning:

\paragraph{Vanilla Instruction Tuning}
The first training stage aims to equip the model with the ability to follow instructions to perform diverse evaluation tasks.
We adopt Flan-T5~\cite{flant5}, an open-source language model as the base model for our evaluator. Based on Flan-T5, we further perform standard instruction tuning on the mixture of four types of tasks: \textit{scoring, comparison, ranking}, and \textit{Boolean QA}, as elaborated in Section~\ref{sec:instruction_tasks}. 

\paragraph{Instruction Tuning with Auxiliary Aspects}

Through our study, we discern that certain evaluation aspects could be interrelated. 
As evidence, in dialogue evaluation~\cite{gopalakrishnan2019topical}
the aspect \texttt{naturalness} usually shows a notable correlation with \texttt{engagingness}. 
When a dialogue response is not \texttt{natural}, it is very likely that human considers the response to be not \texttt{engaging}. While these two aspects are not interchangeable given their different definitions, the evaluation of one aspect can offer useful clues for the evaluation of another potentially related aspect. 
Motivated by this, we enrich our training regimen with an additional instruction tuning stage to leverage potential connections to the target evaluation aspect. 

More precisely, for each instruction-tuning task detailed in Section~\ref{sec:instruction_tasks}, we augment it based on the ground truth evaluation results of a predefined set of auxiliary aspects which are all other aspects collected in the source dataset. 
To convert the evaluation results of auxiliary aspects into natural language that can be fed into the input, we employ a template-based verbalizer, denoted as $v(\cdot)$, which takes in an aspect $a$ and its evaluation score $s$ for an instance, mapping it into a verbalized evaluation $h = v(s, a)$. 
For example, with the aspect \texttt{Consistency} on Data2Text and the evaluation score 0.9 out of 1.0, the verbalized result is phrased as "\textit{This sentence is consistent with the source.}" (see more details in Appendix~\ref{app:xeval}).
We construct the set of verbalized results $\mathcal{H}$ with the verbalizer for each auxiliary aspect (except for the target aspect). This set $\mathcal{H}$ is then concatenated into the additional set of texts in the evaluator's input.
The model then undergoes the second training stage on the instruction tasks enriched with these evaluation results.

\input{algorithm/hint_inference}

\subsection{Inference with Auxiliary Aspects}


At the inference stage, we perform the following steps to evaluate the text on the target aspect: \textbf{First}, we select a set of auxiliary aspects for the target aspect. Based on the definitions of the target aspect and a pool of candidate aspects, we employ Sentence-T5~\cite{ni2022sentence} to encode the definitions and measure the similarity between the sentence embeddings of target aspect definition and each candidate aspect definition. 
We select the aspects with top-$k$ similarity scores as the auxiliary aspects to limit inference cost, where $k$ is a hyperparameter. 
\textbf{Second}, we run an inference process using the Boolean QA task format, where the model predicts either \textit{"Yes"} or \textit{"No"}, as outlined in Section~\ref{sec:instruction_tasks}, on each auxiliary aspect. We convert the prediction into natural language results with the verbalizer.
These verbalized results, denoted as $\mathcal{H}$, are subsequently integrated into the additional set of texts $\mathcal{S}$ for evaluating the target aspect. 
\textbf{Finally}, given the input enhanced by auxiliary aspects, we adopt the same Boolean QA format to compute the evaluation score $c$ for the target aspect:
\begin{align*}
    c = \frac{P(``Yes" | x, \mathcal{S}, a)}{P(``Yes" | x, \mathcal{S}, a) + P(``No" | x, \mathcal{S}, a)}
\end{align*}
where $P(\cdot)$ denotes the probability of the model generating a specific word. 
The pseudo-code of our inference pipeline is in Algorithm~\ref{alg:hint_infer}.





%% file: algorithm/hint_inference.tex
\begin{algorithm}[!t]
\small
\caption{Inference Pipeline}\label{alg:hint_infer}
\KwIn{Set of evaluation aspects $\mathcal{A}$, Target aspect $a_t$, NLG system's output $x$, Additional set of texts $\mathcal{S}$, Scoring function $f(\cdot)$, Evaluation verbalizer $v(\cdot)$, Similarity measure $sim(\cdot)$, Sentence encoder $\mathcal{E}$}
\KwOut{Target score $c_t$}

\tcp{Determine top-$k$ auxiliary aspects}
$L \leftarrow \{ (sim(\mathcal{E}(a), \mathcal{E}(a_t)), a) \mid a \in \mathcal{A} \setminus \{a_t\} \}$

Sort $L$ in descending order based on similarity

$\mathcal{A}^R \leftarrow$ first $k$ aspects from sorted $L$

\tcp{Generate verbalized evaluation results for auxiliary aspects}
Initialize an empty auxiliary evaluation set $\mathcal{H}$

\For{$a_r \in \mathcal{A}^R$}{
    \tcp{Score for auxiliary aspect}
    $c_r \leftarrow  f(x, \mathcal{S}_r, a_r)$ 

    \tcp{Add verbalized evaluation to the auxiliary evaluation set}
    $\mathcal{H} \leftarrow [\mathcal{H}; v(c_r, a_r)]$ 
}

$\mathcal{S}_t \leftarrow [\mathcal{S}_t; \mathcal{H}]$

\tcp{Evaluate the target aspect}
$c_t \leftarrow  f(x, \mathcal{S}_t, a_t)$ 

\textbf{return} $c_t$
\end{algorithm}

%% file: 4experiment.tex
\section{Experiment Setup}

\begin{table*}[ht]
\centering
\resizebox{\textwidth}{!}
{
\begin{tabular}{l|cccccc|c|ccccc|c}

\toprule
\multicolumn{1}{c|}{\multirow{2}[1]{*}{\textbf{Metrics}}} & \multicolumn{7}{c|}{\textbf{Dialogue-level}} & \multicolumn{6}{c}{\textbf{Turn-level}} \\

& \textbf{DEP} & \textbf{LIK} & \textbf{UND} & \textbf{FLE} & \textbf{INF} & \textbf{INQ} & \textbf{{AVG}} & \textbf{INT}   & \textbf{SPE}   & \textbf{COR}   & \textbf{SEM}   & \textbf{UND}    & \textbf{AVG} \\
\midrule 
BARTScore~\cite{yuan2021bartscore}& 0.082 & 0.099 & -0.115 & 0.093 & 0.092 & 0.062 & 0.052 & 0.159          & 0.083          & 0.076          & 0.100          & 0.120          & 0.128  \\
DynaEval~\citep{dynaeval} & 0.498 & 0.416 & 0.365 & 0.383 & 0.426 & 0.410 & 0.416 & 0.327          & 0.346          & 0.242          & 0.202          & 0.200   & 0.263 \\
UniEval~\citep{unieval} & 0.046 & 0.009 & -0.024 & -0.003 & -0.070 & 0.085 & 0.030 & 0.435   & \underline{\textbf{0.381}} & 0.125          & 0.051          & 0.082  &  0.215  \\ 
\midrule
GPTScore (GPT-3-d01)~\citep{gptscore} & \textbf{0.669} & \textbf{0.634} & 0.524 & 0.515 & \textbf{0.602} & 0.503 & \textbf{0.574} & 0.501 & 0.214 & 0.434 & \textbf{0.444} & 0.365 & 0.392 \\ 
GPTScore (GPT-3-d03)~\citep{gptscore} & 0.341 & 0.184 & 0.196 & 0.072 & 0.317 & -0.101 & 0.168 & 0.224 & 0.151 &  0.428 &  0.405 & 0.311 & 0.304   \\
G-Eval (GPT-3.5)\dag~\citep{geval} & 0.339 & 0.392 & 0.123 & 0.344 & 0.232 & 0.101 & 0.259 & 0.30           & 0.280          & 0.430          & 0.390          & 0.274  &  0.335 \\
G-Eval (GPT-4)\dag~\citep{geval} & 0.583 & 0.614 & \textbf{0.602} & \textbf{0.587} & 0.510 & \textbf{0.551} & 0.573  &   \textbf{0.506} & 0.368 & \textbf{0.522} & 0.443 & \textbf{0.438} &  \textbf{0.455} \\
\midrule 

\method{} (Ours) & \underline{0.583} & \underline{0.436} & \underline{0.588} & 0.324 & 0.480 & \underline{0.497} & \underline{0.485} & 0.421 & 0.370 & \underline{0.492} & \underline{0.376} & \underline{0.332} & \underline{0.398} \\
- w/o Training & 0.377 & 0.387 & 0.394 & \underline{0.424} & 0.370 & 0.417 & 0.395 &  0.250 &  0.175 &  0.296 &  0.289 & 
 0.225  &   0.247  \\
- w/o Instructions & 0.350 & 0.333 & 0.495 & 0.355 & 0.425 & 0.435 & 0.399 & \underline{0.477} & 0.353 & 0.203 & 0.255 & 0.211 & 0.300 \\
- w/o Stage-Two Tuning  & 0.388 & 0.324 & 0.555 & 0.384 & \underline{0.582} & 0.437 & 0.445 &  0.372 & 0.282 &  0.418 & 0.329 & 0.311 & 0.342  \\

\bottomrule
\end{tabular}
}
\caption{\textbf{Meta-evaluation on dialogue} based on \textit{unseen} aspects in terms of dialogue-level and turn-level Spearman ($\rho$) correlations on FED. The best overall results are highlighted in \textbf{bold}. We also highlight the best results excluding GPT-based metrics with \underline{underline}. 
\dag: our re-implementation, where we adopt our annotated instructions and aspect definitions as inputs to OpenAI's API to obtain the performance of G-Eval on FED.
}
\label{tab:fed}
\end{table*}

\paragraph{Meta Evaluation}
We meta-evaluate our \method{} on the test split of \evalinstruct, where the details of the test set are introduced as follows. For text summarization, we adopt SummEval \cite{fabbri2021summeval} and QAGS \cite{wang2020asking}. For dialogue generation, we employ Topical-Chat~\cite{gopalakrishnan2019topical} and FED \cite{mehri2020unsupervised}. For data-to-text generation, we utilize SFHOT \& SFRES \cite{wen2015semantically}. \evalinstruct{} contains the following \textit{unseen} aspects: \texttt{topic depth} (DEP), \texttt{likeability} (LIK), \texttt{understandability} (UND), \texttt{flexibility} (FLE), \texttt{informativeness} (INF), \texttt{inquisitiveness} (INQ), \texttt{interestingness} (INT), \texttt{specificity} (SPE), \texttt{correctness} (COR), and \texttt{semantic appropriateness} (SEM).
More detailed descriptions of the test splits, as well as seen and unseen evaluation aspects, are be found in Appendix~\ref{app:baseline}.

\paragraph{Implementation Details} 
We adopt Flan-T5-large (with  \textasciitilde 780M parameters) as our base language model for subsequent finetuning. Without specification, we pick the top-1 aspect during inference, i.e., $k=1$. More implementation details can be found in Appendix~\ref{app:exp}.


\begin{table*}[ht]
\centering
\resizebox{0.9\textwidth}{!}
{
\begin{tabular}{l|cc|cc|cc|cc|cc}

\toprule

\multicolumn{1}{c|}{\multirow{2}[1]{*}{\textbf{Metrics}}} & \multicolumn{2}{c|}{\textbf{Naturalness}}
 & \multicolumn{2}{c|}{\textbf{Coherence}} & \multicolumn{2}{c|}{\textbf{Engagingness}} & \multicolumn{2}{c|}{\textbf{Groundedness}} & \multicolumn{2}{c}{\textbf{AVG}}  \\ 
            & $r$ & $\rho$ & $r$ & $\rho$ & $r$ & $\rho$ & $r$ & $\rho$ & $r$ & $\rho$      \\ \hline

ROUGE-L~\citep{lin2004rouge} &0.176&0.146&0.193&0.203&0.295&0.300&0.310&0.327&0.243&0.244\\
BERTScore~\citep{Zhang2020BERTScore} &0.226&0.209&0.214&0.233&0.317&0.335&0.291&0.317&0.262&0.273\\
USR~\citep{mehri2020usr} & 0.337 & 0.325 & 0.416 & 0.377 & 0.456 & 0.465 & 0.222 & 0.447 & 0.358 & 0.403\\
UniEval~\citep{unieval} & \underline{0.480} & \underline{0.512} & 0.518 & 0.609 & \underline{0.544} & 0.563 & 0.462 & 0.456 & 0.501 & 0.535      \\
\midrule
G-Eval (GPT-3.5)~\citep{geval} &0.532 & 0.539 & 0.519 & 0.544 & \textbf{0.660} & \textbf{0.691} & 0.586 & 0.567 & 0.574 & 0.585     \\
G-Eval (GPT-4)~\citep{geval}  &\textbf{0.549} & \textbf{0.565} & \textbf{0.594} & 0.605 & 0.627 &0.631 &0.531 & 0.551 & \textbf{0.575} & 0.588\\
\midrule
\method{} (Ours) & 0.417 & 0.478 & 0.558 & 0.622 & 0.449 & \underline{0.593} & \underline{\textbf{0.734}} & \underline{\textbf{0.728}} & \underline{0.540} & \underline{\textbf{0.605}}  \\
- w/o Training & 0.054 & 0.051 & 0.063 & 0.073 & 0.258 & 0.298 & 0.427 & 0.436 & 0.200 & 0.214 \\ 
- w/o Instructions &  0.415 & 0.452 & \underline{0.560} & 0.574 & 0.397 & 0.532 & 0.690 & 0.701 & 0.515 & 0.565 \\
- w/o Stage-Two Tuning & 0.396 & 0.446 & 0.581 & \underline{\textbf{0.642}} & 0.408 & 0.569 & 0.725 & 0.706 & 0.528 & 0.592 \\

\bottomrule

\end{tabular}
}
\caption{Turn-level Pearson ($r$) and Spearman ($\rho$) correlations on \textit{seen} aspects on Topical-Chat. The best overall results are highlighted in \textbf{bold}. We also highlight the best results excluding GPT-based metrics with \underline{underline}. }
\label{tab:topical_chat}
\end{table*}

\begin{table*}[tbp]

\centering
\resizebox{0.9\textwidth}{!}
{
\begin{tabular}{l|cc|cc|cc|cc|cc}

\toprule

\multicolumn{1}{c|}{\multirow{2}[1]{*}{\textbf{Metrics}}} & \multicolumn{2}{c|}{\textbf{Coherence}}
 & \multicolumn{2}{c|}{\textbf{Consistency}} & \multicolumn{2}{c|}{\textbf{Fluency}} & \multicolumn{2}{c|}{\textbf{Relevance}} & \multicolumn{2}{c}{\textbf{AVG}}  \\ 
            & $\rho$         & $\tau$         & $\rho$          & $\tau$         & $\rho$         & $\tau$         & $\rho$         & $\tau$         & $\rho$         & $\tau$         \\ 
\midrule
ROUGE-L~\citep{lin2004rouge}     & 0.128          & 0.099          & 0.115           & 0.092          & 0.105          & 0.084          & 0.311          & 0.237          & 0.165          & 0.128          \\ 
MOVERSscore~\cite{zhao2019moverscore}  & 0.159          & 0.118          & 0.157           & 0.127          & 0.129          & 0.105          & 0.318          & 0.244          & 0.191          & 0.148          \\
BERTScore~\cite{Zhang2020BERTScore}   & 0.284          & 0.211          & 0.110           & 0.090          & 0.193          & 0.158          & 0.312          & 0.243          & 0.225          & 0.175          \\
BARTScore~\cite{yuan2021bartscore} & 0.448 & 0.342 & 0.382 & 0.315 & 0.356 & 0.292 & 0.356 & 0.273 & 0.385 & 0.305 \\
UniEval~\cite{unieval} & 0.495 & 0.374 & \underline{0.435} & \underline{0.365} & 0.419 & 0.346 & 0.424 & 0.327& 0.443 & 0.353 \\
\midrule 
GPTScore~\cite{gptscore} &0.434&--&0.449&--&0.403&--&0.381&--&0.417&--
\\
G-Eval (GPT-3.5)~\citep{geval} & 0.440          & 0.335          & 0.386           & 0.318          & 0.424          & 0.347          & 0.385          & 0.293          & 0.401          & 0.320          \\
G-Eval (GPT-4)~\citep{geval}   & \textbf{0.582} & \textbf{0.457} & \textbf{0.507}  & \textbf{0.425} & 0.455 & \textbf{0.378} & \textbf{0.547} & \textbf{0.433} & \textbf{0.514} & \textbf{0.418}\\
\midrule 
\method{} (Ours) & 0.530 & 0.382 & 0.428 & 0.340 & \underline{\textbf{0.461}} & \underline{0.365} & 0.500 & 0.361 & \underline{0.480} & \underline{0.362}  \\
- w/o Training & 0.187 & 0.131 & 0.193 & 0.152 & 0.135 & 0.104 & 0.444 & 0.325 & 0.240 &  0.178 \\ 
- w/o Instructions & 0.458 & 0.333 & 0.414 & 0.328 & 0.395 & 0.309 & 0.496 & 0.359 & 0.441 & 0.333  \\
- w/o Stage-Two Tuning & \underline{0.536} & \underline{0.385} & 0.413 & 0.326 & 0.455 & 0.360 & \underline{0.503} & \underline{0.363} & 0.476 & 0.359 \\

\bottomrule

\end{tabular}
}
\caption{Summary-level Spearman ($\rho$) and Kendall-Tau ($\tau$) correlations of different metrics on SummEval. All aspects are \textit{seen} aspects. The best overall results are highlighted in \textbf{bold}. We also highlight the best results excluding GPT-based metrics with \underline{underline}. 
}
\label{tab:summ}
\end{table*}




\paragraph{Baselines}
We compare our \method{} with the following state-of-the-art NLG evaluation metrics: 
\textbf{(1) UniEval}~\cite{unieval} is a unified multi-aspect evaluator that re-frames the evaluation process as a Boolean QA task; 
\textbf{(2) GPTScore}~\cite{gptscore} is a multi-faceted and training-free evaluation framework that utilizes the output probabilities from LLMs to score generated texts; 
\textbf{(3) G-Eval}~\cite{geval} proposes to leverage large language models such as GPT-3.5 or GPT-4 to assess the text quality with form-filling paradigm in a training-free manner; \textbf{(4) ROUGE-L}~\cite{lin2004rouge}; \textbf{(5) DynaEval}~\cite{dynaeval}; \textbf{(6) BERTScore}~\cite{Zhang2020BERTScore}; \textbf{(7) MoverScore}~\cite{zhao2019moverscore}; \textbf{(8) USR}~\cite{mehri2020usr}; \textbf{(9) BARTScore}~\cite{yuan2021bartscore}. We include more details of baselines (4)-(9) in Appendix~\ref{app:exp} due to space limit.

\paragraph{Variants of \method{}}
We design several variants of \method{} for ablation studies: \textbf{(1) \method{} w/o Training} denotes the original Flan-T5 (without any further finetuning on our proposed \evalinstruct{}); \textbf{(2) \method{} w/o Instructions}: based on Flan-T5, we only conduct prompt-based multi-task training and inference in the same way as~\cite{unieval} where we ask the model to answer Boolean questions without using aspect definitions;
\textbf{(3) \method{} w/o Stage-Two Tuning}: for this variant, we only conduct vanilla instruction tuning in Stage 1 based on Flan-T5. During inference, we directly perform evaluation based on the instructions without using auxiliary aspects.


%% file: 5discussion.tex
\section{Main Results}

\begin{table}[t]
\center 
\resizebox{0.5\textwidth}{!}
{
\begin{tabular}{l|cc|cc|c}
\toprule
\multicolumn{1}{l}{\multirow{2}[1]{*}{\textbf{Metrics}}} & \multicolumn{2}{c}{\textbf{SFRES}} & 
\multicolumn{2}{c}{\textbf{SFHOT}} &
\multicolumn{1}{c}{\multirow{2}[1]{*}{\textbf{AVG}}} \\
 & \textbf{NAT} & \textbf{INFO} & \textbf{NAT} & \textbf{INFO} \\

\midrule

ROUGE-L & 0.169 & 0.103 & 0.186 & 0.110 & 0.142 \\
BERTScore & 0.219 & 0.156 & 0.178 & 0.135 & 0.172 \\
MOVERScore & 0.190 & 0.153 & 0.242 & 0.172 & 0.189 \\
BARTScore & 0.289 & 0.238 & 0.288 & 0.235 &  0.263 \\
UniEval (Summ) & 0.333 & 0.225 & 0.320 & 0.249 & 0.282 \\
\midrule

GPTScore & 0.190 & 0.232 & 0.036 & 0.184 & 0.161  \\
G-Eval (GPT-3.5)\dag &  0.144 & 0.118 & 0.072 & 0.102 & 0.109 \\
G-Eval (GPT-4)\dag  &  \textbf{0.351} & 0.189 & \textbf{0.338} & 0.198 &  0.269  \\

\midrule
\method{} (Ours) & 0.316 & \textbf{0.265} & 0.322 & \textbf{0.310} & \textbf{0.303}  \\
- w/o Training & 0.240 & 0.192 & 0.207 & 0.262 & 0.225 \\
- w/o Instructions & 0.303 & 0.255 & 0.297 & 0.277 & 0.283 \\
- w/o Stage-Two Tuning & 0.322 & 0.257 & 0.311 & 0.292 & 0.295   \\

\bottomrule
\end{tabular}
}
\caption{Spearman correlation on the data-to-text NLG task. NAT and INFO indicate \texttt{Naturalness} and \texttt{Informativeness}, respectively. The best results are highlighted in \textbf{bold}. \dag: our re-implementation.}
\label{tab:d2t}
\end{table}

We report the main results of dialogue evaluation in Table~\ref{tab:fed} and Table~\ref{tab:topical_chat}, summarization in Table~\ref{tab:summ} and Table~\ref{tab:qags} , and data-to-text in Table~\ref{tab:d2t}. Each table is divided into three sections: the top section delineates the performance of traditional metrics and evaluators based on lightweight language models. The middle section shows the performance of the evaluators based on GPTs~\cite{GPT3,OpenAI2023GPT4TR} that are proprietary and much larger than our approach. The bottom section shows the performance of \method{} and its variants. 


\paragraph{Results of Dialogue Evaluation on FED}

To assess \method{}'s ability to generalize to \textit{unseen} aspects, we present the Spearman correlation on FED in Table~\ref{tab:fed}. 
\method{} surpasses the baselines in the top section.
Also, \method{} matches the performance of GPT-based baselines with much fewer parameters. 
The bottom section of the table highlights the improvement achieved by two-stage tuning, incorporating instructions, and integrating auxiliary aspects.
It is worth noting that UniEval achieves notably poor performance on dialogue-level evaluation on FED, which is probably due to UniEval being overfitted to turn-level evaluation and failing to generalize to dialogue-level evaluation.

\paragraph{Results of Dialogue Evaluation on Topical-Chat} 
We also evaluate the performance for the \textit{seen} aspects on Topical-Chat and report the results in Table~\ref{tab:topical_chat}. 
Notably, in addition to the superior performance over lightweight baselines, \method{} also surpasses all GPT-based metrics in averaged Spearman correlation. 
We notice that the correlation of \method{} on \texttt{groundedness} is notably higher than other baselines. 
One plausible reason is that Flan-T5 has been finetuned on related tasks such as natural language inference~\cite{flant5}, as \method{} w/o Training has achieved decent performance without finetuning on \evalinstruct{}.  

\paragraph{Results of Summarization Evaluation}
We use summary-level Spearman and Kendall-Tau correlation to assess various evaluators on SummEval. 
Note that all the aspects in SummEval are seen aspects.
From Table~\ref{tab:summ}, \method{} surpasses lightweight evaluators in averaged Spearman correlation and outperforms both GPTScore and G-Eval (GPT-3.5). G-Eval (GPT-4) consistently excels across all aspects. We speculate this may stem from GPT-4's strong ability to handle long input contexts. 
In addition, we report the results on QAGS in Table~\ref{tab:qags} in Appendix due to the space limit.

\paragraph{Results of Unseen NLG Task Evaluation}
In this experiment, we evaluate \method{} on the unseen data-to-text generation task. 
Table \ref{tab:d2t} shows that while \method{} experiences a slight performance loss in \texttt{naturalness} compared to G-Eval (GPT-4), it consistently excels over all other baselines across all aspects. This underscores the generalization capability of \method{} on unseen NLG tasks. 

\begin{table}[t]
\center 
\resizebox{0.5\textwidth}{!}
{
\begin{tabular}{l|c|c|c|c|c}
\toprule
\textbf{Metrics} & \textbf{Topic.} & \textbf{FED} &  \textbf{Summ.} & \textbf{D2T} & \textbf{AVG} \\ 

\midrule
\method{} (w/o STT) & \textbf{0.592} & \textbf{0.375} & \textbf{0.480} & 0.295 & \textbf{0.436} \\
- w/o Scoring & 0.547 &  0.281 & 0.438 & \textbf{0.300} & 0.392 \\
- w/o Comparison & 0.554 & 0.347 &  0.448 & 0.293 & 0.411 \\
- w/o Ranking & 0.591 & 0.354 & 0.433 & 0.252 & 0.408 \\
- w/o QA & 0.579 & 0.357 & 0.418 & 0.284 & 0.410    \\

\bottomrule
\end{tabular}
}
\caption{Ablation study on stage one instruction tuning task type (Spearman correlation). "w/o STT" denotes the model does not use Stage-Two Tuning. The best results are highlighted in \textbf{bold}. 
}
\label{tab:task_ablation}
\end{table}

\begin{figure*}[htbp]
	\centering
	\includegraphics[width=\linewidth, trim={0 0 0 0},clip ]{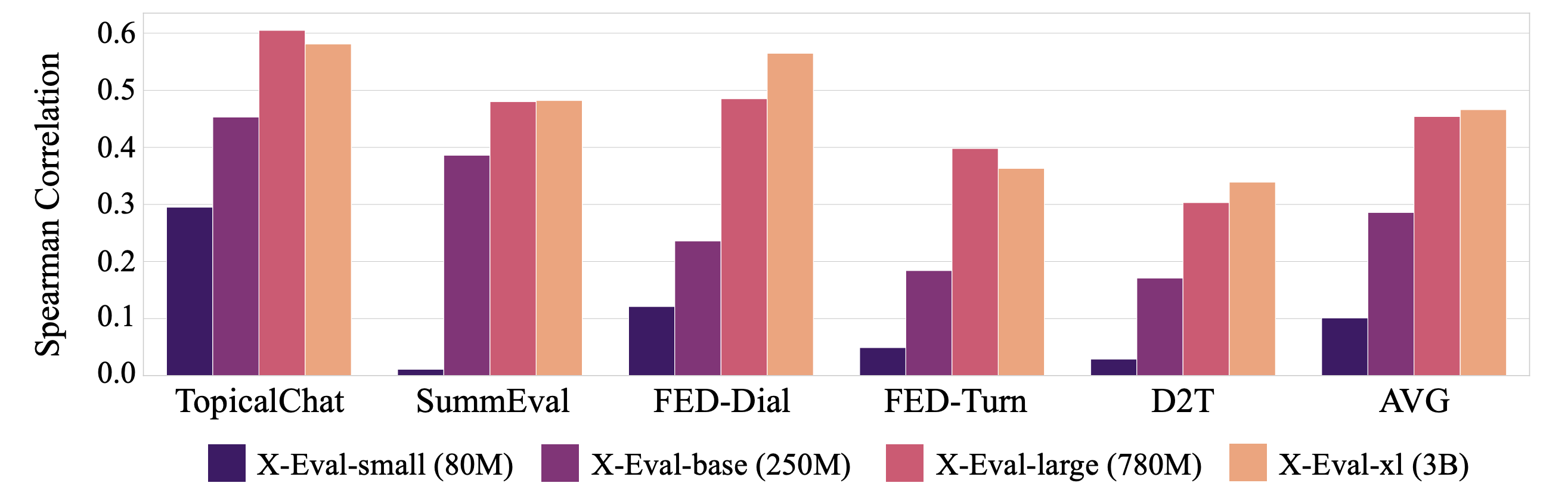}
	\caption{Effect of the scale of language model backbones. For each meta-evaluation benchmark, we report the average Spearman correlation on all the aspects. \method-large (780M) is the default backbone language model throughout all the experiments if there is no specification.
 }
\label{effect_of_scale}
\end{figure*}

\section{Discussions}

\paragraph{Ablation Study of Instruction Tuning Tasks}

We conduct ablation studies to investigate the contribution of incorporating diverse forms of evaluation tasks during instruction tuning. Table \ref{tab:task_ablation} shows the averaged Spearman correlation on each meta-evaluation dataset. 
In general, \method{} trained on the combination of all forms of evaluation tasks, including \textit{scoring, comparison, ranking}, achieves the highest averaged correlation for nearly all tasks. 

\paragraph{Effect of the Scale of Language Model Backbones} We adopt the same training and inference pipelines for the backbones with different scales to show the effect of the models’ size and justify the use of Flan-T5-large. Specifically, we additionally experiment with Flan-T5-small (80M), Flan-T5-base (250M), and Flan-T5-xl (3B) as the backbone models, and term our \method{} respectively. The results are shown in Figure~\ref{effect_of_scale}. From Figure~\ref{effect_of_scale}, the evaluators’ performance consistently increases as the model size increases in general. However, when we upgrade the backbone model from Flan-T5-large to Flan-T5-xl, the performance improvement becomes less significant. Given the trade-off between efficiency and performance, we select Flan-T5-large as the default backbone model of \method{} in our experiments. We include a more detailed performance analysis of the effect of language model backbones in Appendix~\ref{app:exp}.

\begin{table}[t]
\center 
\resizebox{0.49\textwidth}{!}
{
\begin{tabular}{l|c|c|c|c|c}
\toprule
\textbf{Metrics} & \textbf{NAT} & \textbf{COH} & \textbf{ENG} & \textbf{GRO} & \textbf{AVG} \\ 

\midrule

\method{} & \underline{0.478} & 0.622 & \underline{0.593} & \underline{0.728} & \underline{0.605} \\
- Inference w/o Auxiliary Aspects &  0.462 & \underline{0.641} & 0.577 & 0.723 & 0.600 \\
- w/ GT RAA (Upperbound) & \textbf{0.552} & \textbf{0.651} & \textbf{0.703} & \textbf{0.751} &  \textbf{0.664} \\
- w/ Random RAA (Lowerbound) & 0.468 & 0.601 & 0.561 & 0.628 & 0.564 \\

\bottomrule
\end{tabular}
}
\caption{Analysis of error propagation in auxiliary aspects on Topical-Chat in terms of Spearman correlation. We highlight the best results in \textbf{bold} and the best results without using ground truths with \underline{underline}. ``RAA'' denotes the evaluation Results on Auxiliary Aspects.}
\label{tab:error}
\end{table}

\paragraph{Error Propagation from Auxiliary Aspects during Inference}
During inference, \method{} may predict inaccurate evaluations for auxiliary aspects. To investigate their impact, we tailor several baselines: (1) directly applying the model after two-stage tuning to evaluate without auxiliary aspects; (2) using the ground truth (``GT'') evaluation results instead of predicted results for auxiliary aspects (upperbound), and; (3) using random evaluation results for auxiliary aspects (lowerbound).
From Table~\ref{tab:error}, removing auxiliary aspects makes the overall performance drop. The variant with GT results gains improvement in all aspects, which indicates the error in the evaluation of auxiliary aspects does impact the performance of target aspects, but not to a large degree.
Using random results, on the other hand, deteriorates the performance significantly.

\paragraph{Effect of Hyperparameter $k$}
We examine the choice of $k$ in selecting top-$k$ auxiliary aspects during inference. Table \ref{tab:effect_of_k} shows that inference with the top-$1$ auxiliary aspect generally achieves better correlation. We speculate that this may stem from the error propagation during inference on auxiliary aspects, where using more auxiliary aspects potentially introduces more inaccuracies, offsetting their potential performance benefits.

\begin{figure}[tbp]
	\centering
	\includegraphics[width=\linewidth, trim={0.2cm 0.28cm 0.35cm 0.25cm},clip ]{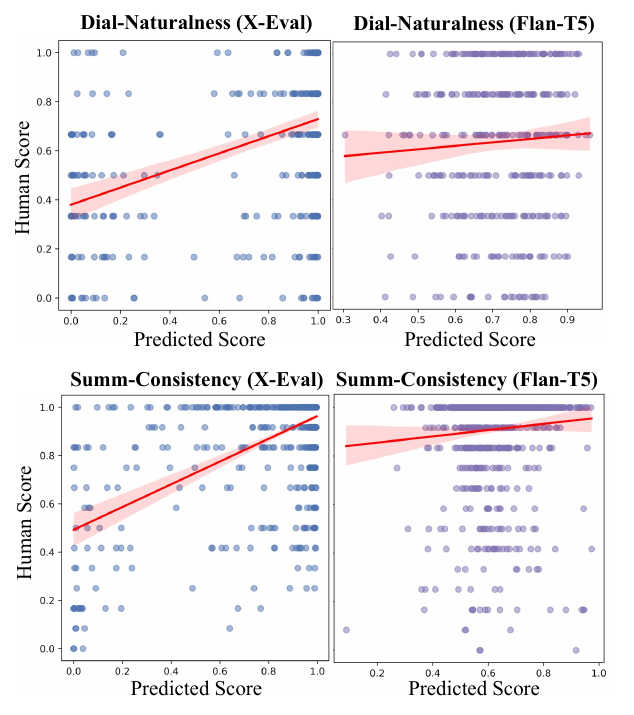}
	\caption{The scatter plots of correlation between human scores and predicted scores of \method{} and Flan-T5, respectively.
 }
\label{fig:corr_graph}
\end{figure}

\paragraph{Qualitative Correlation Analysis on Instruction Tuning} To further investigate the effect of instruction tuning, in Figure~\ref{fig:corr_graph}, we visualize the correlation of our \method{} and Flan-T5 (i.e., ``\method{} w/o Training'') based on \texttt{naturalness} on Topical-Chat and \texttt{consistency} on SummEval. The red lines are linear regression fits to show how well the predicted scores correlate to human judgments linearly. Before instruction tuning, 
the predicted scores are more uniformly distributed regardless of ground truth scores, which results in poor correlation. On the contrary, our \method{} can predict scores that not only achieve better correlation but also are more distinctive (either close to 1 or 0), showing the effectiveness of our instruction tuning.

\begin{table}[t]
\center 
\resizebox{0.45\textwidth}{!}
{
\begin{tabular}{l|c|c|c|c|c}
\toprule
\textbf{Selection} & \textbf{Topic.} & \textbf{FED} &  \textbf{Summ.} & \textbf{D2T} & \textbf{AVG}\\ 

\midrule
Top-1 & \textbf{0.605} & 0.434 & \textbf{0.480} & \textbf{0.303} & \textbf{0.456} \\ 
Top-3 & 0.602 & 0.414 & 0.466 & 0.278 & 0.440 \\
Top-5 & 0.598 & \textbf{0.435} & 0.463 & 0.275 & 0.443 \\

\bottomrule
\end{tabular}
}
\caption{Effect of different $k$ in selecting auxiliary aspect in terms of averaged Spearman correlation. The best results are highlighted in \textbf{bold}. }
\label{tab:effect_of_k}
\end{table}

\paragraph{Visualization of Auxiliary Aspect Selection}
In Figure~\ref{fig:aspect_selection}, we also report the cosine similarity between the sentence embeddings of the aspect definitions used in turn-level dialogue evaluation as the qualitative analysis of our aspect selection strategy. In general, our strategy can select semantically related aspects for target-aspect evaluation.

\paragraph{Analysis of Auxiliary Aspect Selection Strategy}
We also experimented to compare the performance of selecting auxiliary aspects based on seen, unseen, or all aspects, as well as randomly selecting aspects regardless of the definitions. We set the number of auxiliary aspects to 1 in this experiment. From Table~\ref{tab:aspect_pool}, selecting the auxiliary aspect based on all the aspects achieves the best overall performance. Also, we observe a substantial performance degradation when the auxiliary aspect is randomly selected, which shows the effectiveness of our aspect selection strategy.

\input{figures/aspect_selection}

\begin{table}[t]
\center 
\resizebox{0.4\textwidth}{!}
{
\begin{tabular}{l|c|c|c}
\toprule
\textbf{Selection} & \textbf{Topic-Chat} & \textbf{FED-Turn} & \textbf{AVG}\\ 

\midrule
All & 0.605 & 0.398 & \textbf{0.502} \\
Seen & 0.602 & \textbf{0.399} & 0.489 \\
Unseen & \textbf{0.608} & 0.379 & 0.481 \\
Random & 0.592 & 0.381 & 0.475\\

\bottomrule
\end{tabular}
}
\caption{Comparison of different pools of candidate auxiliary aspects in terms of averaged Spearman correlation for turn-level dialogue evaluation. The best results are highlighted in \textbf{bold}.
}
\label{tab:aspect_pool}
\end{table}








%% file: figures/aspect_selection.tex
\begin{figure}[!t]
	\centering
	\includegraphics[width=\linewidth, trim={0 0 0 0},clip ]{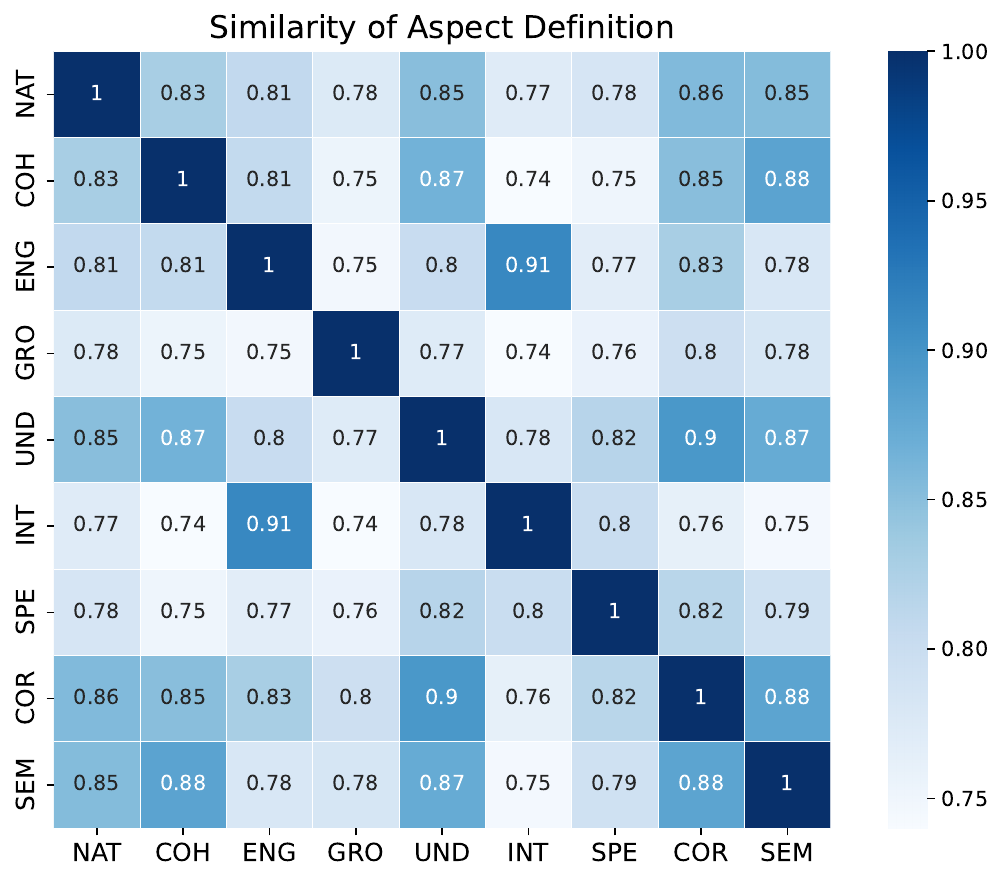}
	\caption{Cosine similarity scores of the sentence embeddings of aspect definition in turn-level dialogue evaluation. \texttt{Naturalness} (NAT), \texttt{coherence} (COH), \texttt{engagingness} (ENG), and \texttt{groundedness} (GRO) are seen aspects, while the rest are unseen aspects. }
\label{fig:aspect_selection}
\end{figure}

%% file: 6conclusion.tex
\section{Conclusion}
In this work, we present \method{}, a novel two-stage instruction-tuning framework for text evaluation across both seen and unseen aspects.
To facilitate training, we collect \evalinstruct{}, the first instruction-tuning dataset for multi-aspect evaluation. 
Extensive experiments on meta-evaluation benchmarks demonstrate that with significantly fewer parameters, \method{} achieves a comparable if not higher correlation with human judgments compared to the state-of-the-art NLG evaluators.

%% file: 7limitation.tex
\section{Limitations}
\paragraph{Limitation of Data Collection}
In this work, we mainly target evaluation tasks in English. Future work can explore evaluation tasks in a more diverse language setting and augment our  \evalinstruct{} dataset. In addition, our dataset focuses on a limited subset of NLG tasks including dialogue, summarization, and data2text. More NLG tasks can be considered in the future. 
\paragraph{Inference Efficiency}
Our algorithm may require multiple rounds of predictions to generate evaluation results from auxiliary aspects in the inference time. While this process imposes additional computational costs, given that the backbone we used is lightweight (with 780M parameters) and efficient, our approach is still significantly more efficient than the evaluators that are much larger, e.g., GPT-4. We leave exploring more efficient inference strategies for future work.

\paragraph{Error Propagation}
During inference, the evaluation results of auxiliary aspects may contain some errors. The errors may affect the final evaluation of the target aspect. 
We leave developing more robust inference algorithms to address the error propagation problem for future works.

%% file: appendix.tex
\section{More Details on \evalinstruct}
\label{app:eval_instrcut}


\begin{figure*}[!htbp]
	\centering
	\includegraphics[width=\linewidth, trim={0 1.5cm 0 0},clip ]{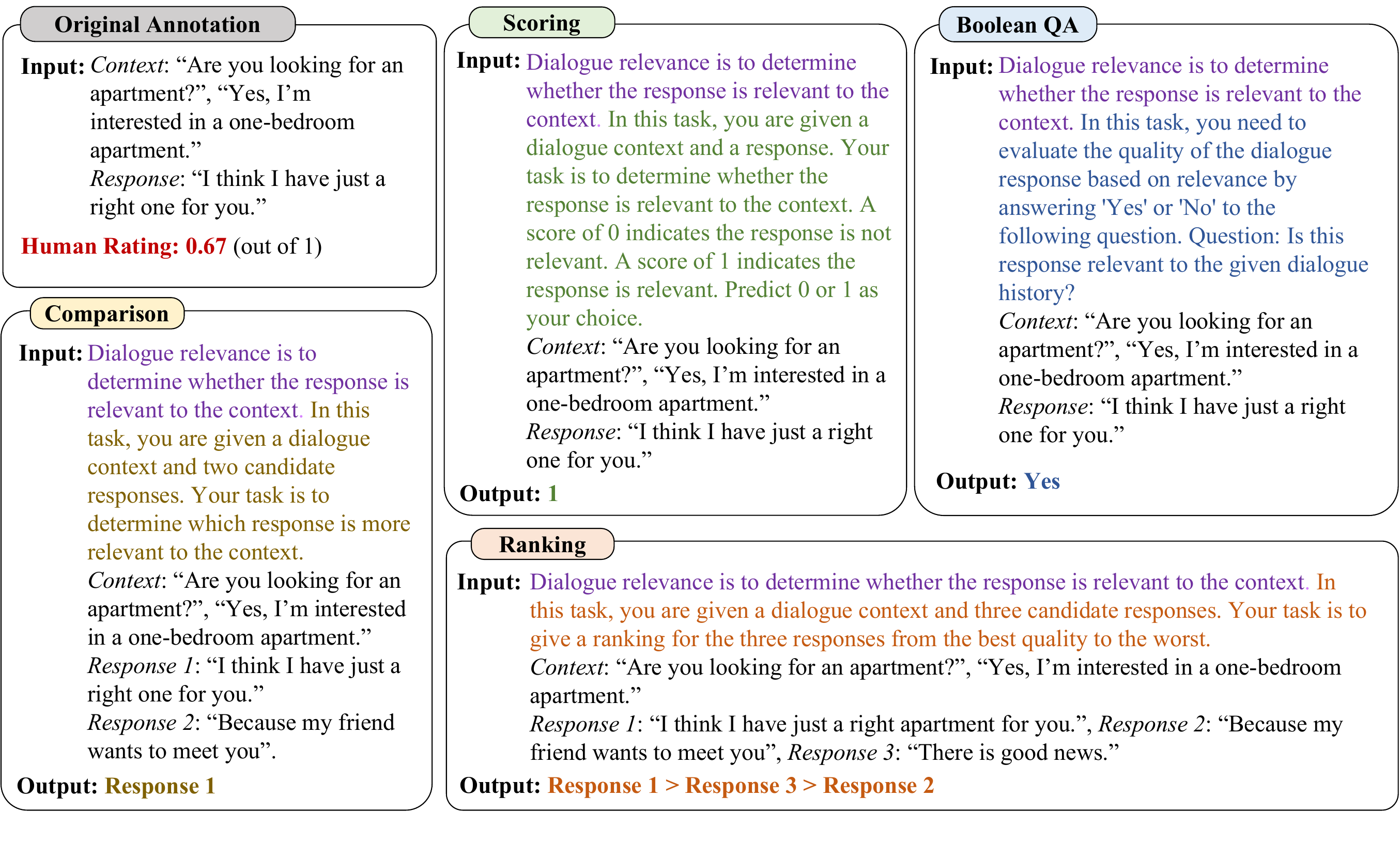}
	\caption{An illustrative example of augmented instruction-tuning tasks from the original annotation. The definition of the aspect is highlighted in purple. The annotated task instructions and the constructed output labels are highlighted in the corresponding colors for each task.
 }
\label{fig_detailed_example}
\end{figure*}

\subsection{Annotation Protocol of Instructions}
\label{app:annotation}
We depict the annotation process for the instructions in \evalinstruct{} as follows. 
To curate the definition for each aspect, we first refer to the definition of the aspect in the original annotation guideline. 
When a definition is absent from the guideline, three human annotators (graduate students studying in computational linguistics or natural language processing areas) construct and revise the definition until they reach an agreement.
The task descriptions and evaluation protocols are also written by three human annotators in similar annotation protocols.

\subsection{Augmenting Instruction-tuning Tasks }
\label{app:aug_it_task}
We show the seen aspects, their corresponding source datasets where we collect the training data, constructed tasks, and the number of training instances for each task in Table~\ref{tab:summ_task_list} and Table~\ref{tab:dial_task_list}. 
For the way we count the number of aspects, we treat the aspects with the same name but in different NLG tasks as different aspects. For example, the \textit{naturalness} aspect in dialogue evaluation and data2text evaluation are considered different under these two settings, although they have the same aspect name.
More specifically, in our \evalinstruct{} dataset, \textit{understandability} is counted twice for dialogue-level and turn-level dialogue evaluation; \textit{naturalness} is counted twice for turn-level dialogue evaluation and data-to-text evaluation; \textit{informativeness} is counted twice for dialogue-level dialogue evaluation and data-to-text evaluation.
We also include an example of how we augment instruction-tuning tasks from the original annotation in Figure~\ref{fig_detailed_example}.

\subsection{Aspect Definition}
\label{app:asp_def}
We present the annotated definitions in \evalinstruct{} in the following. We show the definitions of seen aspects on dialogue evaluation on Table~\ref{tab:dial_def_seen}, unseen aspects on dialogue evaluation on Table~\ref{tab:dial_def_unseen}, and the aspects on summarization on Table~\ref{tab:summ_def}.

\begin{table}[!t]
\center 
\resizebox{0.5\textwidth}{!}
{
\begin{tabular}{l|c|c|c}
\toprule
\textbf{Metrics} & \textbf{CNN} & 
\textbf{XSUM} &
\textbf{AVG} \\

\midrule

ROUGE-L~\cite{lin2004rouge} & 0.324 & -0.011 & 0.156 \\
BERTScore~\cite{Zhang2020BERTScore} & 0.505 & 0.008 & 0.256 \\
MOVERScore~\cite{zhao2019moverscore} & 0.347 & 0.044 & 0.195 \\
BARTScore~\cite{yuan2021bartscore} & \underline{0.680} & 0.159 & 0.420 \\
UniEval~\cite{unieval} & 0.662 & 0.488 & 0.575  \\
\midrule

GPTScore~\cite{gptscore} &  0.649 & 0.238 & 0.443 \\
G-Eval (GPT-3.5)~\cite{geval} & 0.516 & 0.406 & 0.461 \\
G-Eval (GPT-4)~\cite{geval}  &  \textbf{0.685} & \textbf{0.537} & \textbf{0.611} \\

\midrule
\method{} (Ours) &  0.656 & \underline{0.500} & \underline{0.578} \\
\bottomrule
\end{tabular}
}
\caption{Spearman correlation on the summarization task based on the \texttt{consistency} aspect on QAGS. The best results are highlighted in \textbf{bold}.  We also highlight the best results among lightweight (with <7B parameters) and open-source metrics with \underline{underline}.}
\label{tab:qags}
\end{table}

\subsection{Source Datasets for Meta Evaluation}
\label{app:baseline}

\paragraph{SummEval \cite{fabbri2021summeval}} is an evaluation benchmark for summarization which contains human ratings of 100 summaries along four evaluation dimensions: \texttt{fluency}, \texttt{coherence}, \texttt{consistency}, and \texttt{relevance}.

\paragraph{QAGS \cite{wang2020asking}} is a benchmark for identifying and evaluating hallucinations in the summarization task. It aims to measure the factual \texttt{inconsistencies} of generated summaries.

\paragraph{Topical-Chat~\cite{gopalakrishnan2019topical}} is a knowledge-grounded human-human conversation dataset. Following \cite{unieval}, we utilize human ratings collected by \cite{mehri2020usr} for Topical-Chat as the benchmark for evaluating dialog response generation. The assessment consider five aspects: \texttt{naturalness}, \texttt{coherence}, \texttt{engagingness}, \texttt{groundedness}, and \texttt{understandability}.

\paragraph{FED \cite{mehri2020unsupervised}} is an evaluation benchmark for fine-grained dialog evaluation. It comprises human annotations evaluated across eighteen dialog aspects at both the turn-level and the dialog-level.

\paragraph{SFHOT \& SFRES \cite{wen2015semantically}} are evaluation benchmarks for data-to-text task. They provide information about restaurants and hotels in San Francisco. The generated text is evaluated based on two aspects: \texttt{informativeness} and \texttt{naturalness}.

\section{More Details on \method{}}
\label{app:xeval}
\paragraph{Pseudo-code of Inference Pipeline} We provide the pseudo-code of our proposed inference pipeline for \method{} in Algorithm~\ref{alg:hint_infer}.

\paragraph{More Details on Verbalizer $v$ and its Templates}
We design a template-based verbalizer to convert the evaluation results of auxiliary aspects into natural language evaluation that can be integrated into the instructions. 
More formally, the inputs of the verbalizer $v$ contain aspect $a$ and evaluation score $s$ (in the range of 0-1). We first adopt a threshold $\delta$ (we set $\delta=0.5$ throughout all experiments) to get a \textit{binary} label that indicates the quality is \textit{"positive"} (if $s>\delta$) or \textit{"negative"} (if $s\leq\delta$). Given this label and the aspect $a$, we map the results into a template in natural language accordingly. The verbalized results will then be integrated into the instructions.
We construct the templates for each aspect by deriving from aspect definition. We apply the annotation protocol that three human annotators revise the templates together until they reach a consensus.
We show the verbalized templates in Table~\ref{tab:dial_verbalize_template} for dialogue evaluation and Table~\ref{tab:summ_verbalize_template} for summarization evaluation.

\section{More Details on Experiments}
\label{app:exp}
\paragraph{More Implementation Details} We use the checkpoint released on HuggingFace for \texttt{Flan-T5-large}\footnote{\url{https://huggingface.co/google/flan-t5-large}}.
In the first training stage, we set the number of epochs to 2, the learning rate to 5e-05, and the maximum source length to 1024. The second training stage shares the same setup except the number of epochs set to 1. We set the maximum source length during inference to 2048 and pick the top-1 aspect during inference, i.e., $k=1$. We use \texttt{sentence-T5-large}\footnote{\url{https://huggingface.co/sentence-transformers/sentence-t5-large}} to compute the embeddings for aspect definition for auxiliary aspect selection. All the experiments are conducted on NVIDIA A40 GPUs including both training and inference.

\paragraph{More Details on Baselines}
We include more details for the following baselines that are omitted in the main paper due to page limit:
\textbf{(4) ROUGE-L}~\cite{lin2004rouge} counts the overlap (i.e., longest common subsequence) between the text to be evaluated and reference to indicate text quality;
\textbf{(5) DynaEval}~\cite{dynaeval} adopts a graph convolutional network to model dialogue's structure to facilitate evaluation;
\textbf{(6) BERTScore}~\cite{Zhang2020BERTScore} is a similarity-based evaluator. It uses the contextualized representation from BERT~\cite{devlin2019bert} to compute the similarity between the generated text and reference;
\textbf{(7) MoverScore}~\cite{zhao2019moverscore} goes beyond BERTScore by utilizing soft alignments and new aggregation methods on the layer-wise information;
\textbf{(8) USR}~\cite{mehri2020usr} is an unsupervised and reference-free evaluation metric to measure multiple desirable qualities of dialog; 
\textbf{(9) BARTScore}~\cite{yuan2021bartscore} is a unified evaluator based on BART~\cite{lewis2019bart}, which uses the average likelihood of the model output as the metric. Note that for all single-aspect metrics, we compute the correlation between the single predicted evaluation and the human rating of each fine-grained aspect, respectively.

\begin{table*}[t]
\center 
\resizebox{\textwidth}{!}
{
\begin{tabular}{l|c|c|c|c|c|c|c}
\toprule
\textbf{Model} & \textbf{\# Parameters} & \textbf{TopicalChat} & \textbf{SummEval} & \textbf{FED-Dialog} & \textbf{FED-Turn} &\textbf{Data2Text} & \textbf{AVG} \\ 

\midrule

\method-large (Default Ver.)    & 780M   & 0.605   & 0.480   & 0.485   & 0.398   & 0.303   & 0.454   \\
\method-LLaMA-LoRA & 7B  & 0.519 & 0.448 & 0.427 & 0.351 & 0.337 & 0.416 \\

\bottomrule
\end{tabular}
}
\caption{Effect of the scale of language model backbones. For each meta-evaluation benchmark, we report the average Spearman correlation on all the aspects. }
\label{tab:effect_of_scale_llama}
\end{table*}

\input{tables/task_list}

\paragraph{More Results on the Effect of the Scale of Language Model Backbones} 

We further conducted an experiment on using another language model backbone. Specifically, we adopt LLaMA-7B-chat~\cite{llama} as the backbone model and adopt LoRA parameter-efficient tuning~\cite{hu2022lora} during the two-stage instruction tuning. We report the performance in Table~\ref{tab:effect_of_scale_llama}. 

\input{tables/dial_aspect_def}

\input{tables/verbalizer_template}

%% file: tables/task_list.tex
\begin{table*}[!ht]
    \centering
    \resizebox{\textwidth}{!}
    {
    \begin{tabular}{l|p{10cm}|c|c}
    \toprule
        \textbf{Aspect} & \textbf{Datasets} & \textbf{Task} & \textbf{\# Instances} \\ 
        \midrule
        \multirow{4}{*}{Accuracy} & \multirow{4}{10cm}{TL;DR~\cite{tldr}} & Scoring &  5,000 \\
        & & Boolean QA &  5,000 \\
        & & Comparison & 898 \\
        & & Ranking & 599 \\
        \midrule
        \multirow{4}{*}{Coherence} & \multirow{4}{10cm}{TL;DR~\cite{tldr}, UniEval~\cite{unieval}} & Scoring & 5,000 \\
        & & Boolean QA & 5,000 \\
        & & Comparison & 734 \\
        & & Ranking & 425\\
        \midrule
        \multirow{4}{*}{Coverage} & \multirow{4}{10cm}{TL;DR~\cite{tldr}} & Scoring & 5,000 \\
        & & Boolean QA & 4,354  \\
        & & Comparison & 1,028 \\
        & & Ranking & 964 \\
        \midrule
        Consistency & UniEval~\cite{unieval} & Boolean QA & 15,000 \\
        \midrule
        Fluency & UniEval~\cite{unieval} & Boolean QA & 15,000 \\
        \midrule
        Relevance & UniEval~\cite{unieval} & Boolean QA & 15,000 \\

        \bottomrule
    \end{tabular}
    }
    \caption{The full list of apects, the corresponding datasets and tasks on summarization evaluation collected in the training split of \evalinstruct{}. }
    \label{tab:summ_task_list}
\end{table*}

\begin{table*}[!ht]
    \centering
    \resizebox{\textwidth}{!}
    {
    \begin{tabular}{l|p{10cm}|c|c}
    \toprule
        \textbf{Aspect} & \textbf{Datasets} & \textbf{Task} & \textbf{\# Instances} \\ 
        \midrule
        \multirow{4}{*}{Relevance} & \multirow{4}{10cm}{DailyDialog++~\cite{dailydialog}} & Scoring &  2,000 \\
        & & Boolean QA & 2,000 \\
        & & Comparison & 2,000 \\
        & & Comparison (w/ NOTA) & 2,000 \\
        \midrule
        \multirow{2}{*}{Coherence} & \multirow{2}{10cm}{HolisticDial~\cite{holisticdial}; DSTC9~\cite{dstc9}; UniEval~\cite{unieval}} & Scoring & 2,400  \\
        & & Boolean QA & 17,200 \\
        \midrule
        \multirow{2}{*}{Consistency} & \multirow{2}{10cm}{DSTC9~\cite{dstc9}} & Scoring & 2,200\\
        & & Boolean QA & 2,200 \\
        \midrule
        \multirow{2}{*}{Diversity} & \multirow{2}{10cm}{DSTC9~\cite{dstc9}} & Scoring & 2,200 \\
        & & Boolean QA & 2,200\\
        \midrule
        Engagingness & UniEval~\cite{unieval} & Boolean QA & 15,000 \\
        \midrule
        Groundedness & UniEval~\cite{unieval} & Boolean QA & 15,000 \\
        \midrule
        Naturalness & UniEval~\cite{unieval} & Boolean QA & 15,000 \\
        \midrule
        Fluency & HolisticDial~\cite{holisticdial} & Scoring & 200 \\
        
        \bottomrule
    \end{tabular}
    }
    \caption{The full list of apects, the corresponding datasets and tasks on dialogue evaluation collected in the training split of \evalinstruct{}. ``NOTA'' indicates the comparison task consists of the case of ``None Of The Above'', where the quality of two candidates is tied. }
    \label{tab:dial_task_list}
\end{table*}

%% file: tables/dial_aspect_def.tex
\begin{table*}[!ht]
    \centering
    \resizebox{\textwidth}{!}
    {
    \begin{tabular}{l|m{14cm}}
    \toprule
        \textbf{Aspect} & \textbf{Definition} \\ \midrule
        Naturalness & Naturalness in dialogue evaluation refers to the degree to which a response in a conversational context mirrors the characteristics, language use, and structure typical of a human conversational partner. \\ \midrule
        Coherence & Coherence refers to the logical and consistent interconnection of utterances and exchanges throughout a conversation. It represents the extent to which a dialogue system maintains relevance, consistency, and meaningful progression within the discourse, ensuring that the flow and structure of the conversation align with expected conversational norms and the ongoing context. \\ \midrule
        Engagingness & Engagingness in the context of dialogue evaluation refers to the degree to which a response fosters continued interaction, maintains or elevates interest, and stimulates a compelling exchange of ideas, emotions, or information between participants. \\ \midrule
        Groundedness & Dialogue groundedness measures how well does the response use the given fact. A response with weak groundedness means the response does not mention or refer to the fact at all. A response with good groundedness means the response uses the fact well. \\ \midrule
        Relevance & Relevance in dialogue evaluation refers to the measure of applicability, pertinence, or connection of a given response to the preceding conversational context and/or the explicitly posed question or statement. \\ \midrule
        Fluency & Fluency in dialogue evaluation refers to the degree of fluidity, coherence, and linguistic correctness in a generated response. It encompasses not only the grammatical and syntactic accuracy but also the seamless flow of ideas, the smooth transition between topics, and the naturalness of the language used, echoing human-like conversation patterns. \\ 
        \bottomrule
    \end{tabular}
    }
    \caption{The full list and definitions of \textit{seen} aspects on dialogue evaluation collected in \evalinstruct{}. }
    \label{tab:dial_def_seen}
\end{table*}

\begin{table*}[!ht]
    \centering
    \resizebox{\textwidth}{!}
    {
    \begin{tabular}{l|m{14cm}}
    \toprule
        \textbf{Aspect} & \textbf{Definition} \\ \midrule
        Topic Depth & Topic depth refers to the ability of a dialogue system to engage in extensive, detailed, and multi-turn discussions on a particular subject. \\ \midrule
        Likeability & Likeability refers to the degree to which an interactive system presents a pleasant, engaging, and affable conversational style that resonates positively with the user. \\ \midrule
        Understandability & Understandability reflects the ability of a conversational system to correctly parse and interpret user inputs, reflect an appropriate comprehension of the context, and generate contextually relevant responses. \\ \midrule
        Flexibility & Flexibility measures the system's capacity to understand and react appropriately to a wide range of conversational scenarios, and not merely those for which it was explicitly programmed or trained. It implies the capacity to engage in a diverse array of topics, offer meaningful responses in unexpected situations, and adjust conversational strategies based on the evolving context or user input. \\ \midrule
        Informativeness & Informativeness refers to the quality and relevance of the information that a dialogue system provides in response to user inputs. It captures the system's ability to offer novel, detailed, accurate, and appropriate information that aligns with the user's requests or needs. \\ \midrule
        Inquisitiveness & Inquisitiveness pertains to the consistent exhibition of the capacity to ask meaningful, contextually appropriate, and well-timed questions within a conversation by a dialogue system. This behavior is exhibited in the pursuit of greater comprehension, clarifying ambiguities, furthering the dialogue, or driving deeper engagement with the conversation partner. \\ \midrule
        Interestingness & Interestingness refers to the degree to which a response stimulates engagement, thought, or emotional reaction in the average user, fostering a desire to continue the conversation or explore the topic further. It is a measure of the response's capacity to capture the user's attention and maintain their engagement over time. \\ \midrule
        Specificity & Specificity measures to what degree the response is unique, personalized, or pertinent to the specific details of the preceding user inputs or dialogue context, as opposed to being generic, universally applicable, or independent of the conversational specifics. \\ \midrule
        Correctness & Correctness in dialogue evaluation measures to the extent to which a generated response correctly reflects, comprehends, and addresses the salient elements, inferences, and implications in the preceding conversation context. \\ \midrule
        Semantic Appropriateness & Semantic appropriateness is the measure of the extent to which a response in a dialogue maintains logical, meaningful, and contextually fitting alignment with the preceding discourse elements, while adhering to the rules and principles of the language used in the conversation. \\ \bottomrule
    \end{tabular}
    }
    \caption{The full list and definitions of \textit{unseen} aspects on dialogue evaluation collected in \evalinstruct{}. }
    \label{tab:dial_def_unseen}
\end{table*}

\begin{table*}[!ht]
    \centering
    \resizebox{\textwidth}{!}
    {
    \begin{tabular}{l|m{14cm}}
    \toprule
        \textbf{Aspect} & \textbf{Definition} \\ \midrule
        Accuracy & The accuracy aspect measures how the factual information in the summary accurately matches the post. A summary is accurate if it doesn’t say things that aren’t in the article, it doesn’t mix up people, and generally is not misleading. If the summary says anything at all that is not mentioned in the post or contradicts something in the post, it should be considered as an inaccurate summary. \\ \midrule
        Coherence & The coherence aspect measures how coherent is the summary on its own. A summary is coherent if, when read by itself, it’s easy to understand and free of English errors. A summary is not coherent if it’s difficult to understand what the summary is trying to say. Generally, it’s more important that the summary is understandable than it being free of grammar errors. \\ \midrule
        Coverage & The coverage aspect measures how well does the summary cover the important information in the post?” A summary has good coverage if it mentions the main information from the post that’s important to understand the situation described in the post. A summary has poor coverage if someone reading only the summary would be missing several important pieces of information about the situation in the post. A summary with good coverage should also match the purpose of the original post (e.g. to ask for advice). \\ \midrule
        Consistency & The consistency aspect measures the factual alignment between the summary and the summarized source. A factually consistent summary contains only statements that are entailed by the source document. You also need to penalize summaries that contained hallucinated facts. \\ \midrule
        Fluency & Fluency measures the quality of individual sentences. A fluent summary should have no formatting problems, capitalization errors or obviously ungrammatical sentences (e.g., fragments, missing components) that make the text difficult to read.  \\ \midrule
        Relevance & Relevance measures the selection of important content from the source. The summary should include only important information from the source document. You should penalize summaries which contain redundancies and excess information. \\
        
        \bottomrule
    \end{tabular}
    }
    \caption{The full list and definitions of aspects of summarization evaluation collected in \evalinstruct{}. }
    \label{tab:summ_def}
\end{table*}

%% file: tables/verbalizer_template.tex
\begin{table*}[!ht]
    \centering
    \resizebox{0.95\textwidth}{!}
    {
    \begin{tabular}{l|m{12cm}}
    \toprule
        \textbf{Aspect} & \textbf{Verbalizer Template} \\ \midrule
        \multirow{2}{*}{Naturalness} & NEG: The response is unnatural. \\
        & POS: The response is natural. \\
        \midrule
        \multirow{2}{*}{Coherence} & NEG: The response drastically changes topic or ignores the conversation history. \\
        & POS: The response is on topic and strongly acknowledges the conversation history. \\
        \midrule
\multirow{2}{*}{Engagingnes} & NEG: The response is generic and dull. \\
        & POS: The response is interesting or presents an interesting fact. \\
        \midrule
\multirow{2}{*}{Groundedness} & NEG: Given the interesting fact that the response is conditioned on, the response does not mention or refer to the fact at all. \\
        & POS: Given the interesting fact that the response is conditioned on, the response uses the fact well. \\
        \midrule
\multirow{2}{*}{Relevance} & NEG: The response is not relevant to the conversation.  \\
        & POS: The response is relevant to the conversation.  \\
        \midrule
\multirow{2}{*}{Fluency} & NEG: The response is not fluently written.  \\
        & POS: The response is fluently written. \\
        \midrule
        \midrule
        \multirow{2}{*}{Topic Depth} & NEG: The system cannot discuss topics in depth.  \\
        & POS: The system is able to discuss topics in depth. \\
        \midrule
\multirow{2}{*}{Likeability} & NEG: The system cannot display a likeable personality.  \\
        & POS: The system is able to display a likeable personality. \\
        \midrule
\multirow{2}{*}{Understandability} & NEG: The response is difficult to understand. You do not know what the person is trying to say.  \\
        & POS: The response is understandable. You know what the person is trying to say. \\
        \midrule
\multirow{2}{*}{Flexibility} & NEG: The system is not flexible and adaptable to the user and their interests.  \\
        & POS: The system is flexible and adaptable to the user and their interests. \\
        \midrule
\multirow{2}{*}{Informativeness} & NEG: The system is not informative throughout the conversation.  \\
        & POS: The system is informative throughout the conversation. \\
        \midrule
\multirow{2}{*}{Inquisitiveness} & NEG: The system is not inquisitive throughout the conversation.  \\
        & POS: The system is inquisitive throughout the conversation.  \\
        \midrule
\multirow{2}{*}{Interestingness} & NEG: To the average person, the response is not interesting.  \\
        & POS: To the average person, the response is interesting. \\
        \midrule
\multirow{2}{*}{Specificity} & NEG: The response is too generic and not specific to the conversation.  \\
        & POS: The response is specific to the conversation.  \\
        \midrule
\multirow{2}{*}{Correctness} & NEG: There was a misunderstanding of the conversation.  \\
        & POS: The response is correct in the context of the conversation.  \\
        \midrule
\multirow{2}{*}{Semantic Appropriateness} & NEG: The response is not semantically appropriate.  \\
        & POS: The response is semantically appropriate. \\
        \bottomrule
    \end{tabular}
    }
    \caption{The full list of verbalizer templates that are used to convert the evaluation results of auxiliary aspects for dialogue evaluation collected in \evalinstruct{}. "POS" and "NEG" indicate \textit{"positive"} and \textit{"negative"}, respectively. }
    \label{tab:dial_verbalize_template}
\end{table*}

\begin{table*}[!ht]
    \centering
    \resizebox{0.8\textwidth}{!}
    {
    \begin{tabular}{l|m{12cm}}
    \toprule
        \textbf{Aspect} & \textbf{Verbalizer Template} \\ \midrule
        \multirow{2}{*}{Accuracy} & NEG: The factual information in the summary cannot accurately match the post. It says things that aren’t in the article, it mixes up people, or generally is misleading. \\
        & POS: The factual information in the summary accurately match the post. It doesn’t say things that aren’t in the article, it doesn’t mix up people, and generally is not misleading. \\
        \midrule
\multirow{2}{*}{Coherence} & NEG: The summary is not coherent as it lacks a logical flow and has disjointed information, making it difficult to understand the main topic or argument. \\
        & POS: The summary is well-structured and well-organized and it is built from sentence to sentence to a coherent body of information about a topic. \\
        \midrule
\multirow{2}{*}{Coverage} & NEG: The summary has poor coverage on the important information in the post, e.g., someone reading only the summary would be missing several important pieces of information about the situation in the post. \\
        & POS: The summary has good coverage since it mentions the main information from the post that’s important to understand the situation described in the post and also match the purpose of the original post. \\
        \midrule
\multirow{2}{*}{Consistency} & NEG: The summary is not factually consistent with the original post as it  introduces factual inaccuracies or hallucinated facts that are not present in or supported by the original source document. \\
        & POS: The summary has good factual alignment between the summary and the summarized source. It contains only statements that are entailed by the source document. \\
        \midrule
\multirow{2}{*}{Fluency} & NEG: The summary is not fluent as it contains formatting problems, capitalization errors or obviously ungrammatical sentences (e.g., fragments, missing components) that make the text difficult to read. \\
        & POS: This is a fluent summary as it generally does not have formatting problems, capitalization errors or obviously ungrammatical sentences (e.g., fragments, missing components) that make the text difficult to read. \\
        \midrule
\multirow{2}{*}{Relevance} & NEG: This summary is not relevant to the source document as it contains redundancies or excess information. \\
        & POS: The summary generally includes relevant content, capturing some key points from the source. \\
        \bottomrule
    \end{tabular}
    }
    \caption{The full list of verbalizer templates that are used to convert the evaluation results of auxiliary aspects for summarization evaluation collected in \evalinstruct{}. "POS" and "NEG" indicate \textit{"positive"} and \textit{"negative"}, respectively.}
    \label{tab:summ_verbalize_template}
\end{table*}